\documentclass[runningheads]{llncs}
\usepackage{graphicx}
\usepackage{hyperref}
\usepackage{cite}
\usepackage{amsmath,amssymb,amsfonts,bm,mathtools}

\usepackage{textcomp}
\usepackage[dvipsnames]{xcolor}
\def\BibTeX{{\rm B\kern-.05em{\sc i\kern-.025em b}\kern-.08em
    T\kern-.1667em\lower.7ex\hbox{E}\kern-.125emX}}
\usepackage{graphicx}  
\usepackage{algorithm,algorithmic}

\usepackage{enumitem}
\usepackage{multirow}
\usepackage{makecell}

\usepackage{lmodern}
\DeclareMathAlphabet{\fol}{OT1}{lmtt}{b}{n}

\newcommand{\thetab}{\bm\theta}
\newcommand{\F}{\mathcal{F}}

\newcommand{\x}{\mathbf{x}}

\usepackage{xcolor}

\spnewtheorem*{example*}{Example}{\bfseries}{\itshape}

\begin{document}
\title{Neural Networks for Relational Data}
%
%
\author{Navdeep Kaur\inst{1} 
\and
Gautam Kunapuli\inst{1}
\and
Saket Joshi\inst{2}
\and 
Kristian Kersting\inst{3}
\and
Sriraam Natarajan\inst{1}
}
\authorrunning{N. Kaur et al.}
%
\institute{The University of Texas at Dallas \\
\email{\{Navdeep.Kaur,  Gautam.Kunapuli,  Sriraam.Natarajan\}@utdallas.edu}
\and
Amazon Inc. \email{\{saketjoshi@gmail.com\}}
 \and
TU Darmstadt, Germany \email{\{kersting@cs.tu-darmstadt.de\}}}
\maketitle              
\vspace{-0.25in}
\begin{abstract}
While deep networks have been enormously successful over the last decade, they rely on flat-feature vector representations, which makes them unsuitable for richly structured domains such as those arising in applications like social network analysis. Such domains rely on relational representations to capture complex relationships between entities and their attributes. Thus, we consider the problem of learning neural networks for relational data. We distinguish ourselves from current approaches that rely on expert hand-coded rules by learning relational random-walk-based features to capture local structural interactions and the resulting network architecture. We further exploit parameter tying of the network weights of the resulting relational neural network, where instances of the same type share parameters. Our experimental results across several standard relational data sets demonstrate the effectiveness of the proposed approach over multiple neural net baselines as well as state-of-the-art statistical relational models.

\end{abstract}

\section{Introduction}

While successful, deep networks have a few important limitations. Apart from the key issue of interpretability, the other major limitation is the requirement of a flat inputs (vectors, matrics, tensors), which limits applications to tabular, {\em propositional representations}. On the other hand, {\em symbolic and structured representations} 
\cite{srlBook,staraiBook,prm,mln06,PSL2017}  have the advantage of being interpretable, while also allowing for rich representations that allow for learning and reasoning with multiple levels of abstraction. This representability allows them to model complex data structures such as graphs far more easily and interpretably than basic propositional representations. While expressive, these models do not incorporate or discover latent relationships between features as effectively as deep networks. 

Consequently, there has been focus on achieving the dream team of logical and statistical learning methods such as {\em relational neural networks}~\cite{KazemiPoole18-RelNNs,SourekEtAl-16-PredRelNNs}. While specific architectures differ, these methods generally employ {\em hand-coded relational rules} or Inductive Logic Programming (ILP, \cite{LavracDzeroski93-ILPBook}) to identify the domain's structural rules; these rules are then used with the observed data to unroll and learn a neural network. We improve upon these methods in two specific ways: (1) we employ a rule learner that has been recently successful to automatically extract {\em interpretable} rules that are then employed as hidden layer of the neural network; (2) we exploit the notion of {\em parameter tying} from the perspective of statistical relational learning models that allow multiple instances of the same rule share the same parameter. These two extensions significantly improve the adaptation of neural networks (NNs) for relational data.

We employ {\em Relational Random Walks}~\cite{pcrw10} to extract relational rules from a database, which are then used as the first layer of the NN. These random walks have the advantages of being learned from data (instead of time-consumingly hand-coded), and interpretable (as walks are rules in a database schema). Given evidence (facts), relational random walks are instantiated (grounded); parameter tying ensures that groundings of the same random walk share the same parameters with far fewer network parameters to be learned during training. 

For combining outputs from different groundings of the {\em same clause}, we employ combination functions~\cite{NatarajanEtAl08-CRs, JaegerICML07}. For instance, given a rule: $\fol{Professor(P)}$, $\fol{\, Author(P,U), \, Author(S,U), \, Student(S)}$, the $\fol{ana}$-$\fol{bob}$ $\fol{Professor}$-$\fol{Student}$ pair could have coauthored $6$ papers, while the $\fol{cam}$-$\fol{dan}$  pair could have coauthored $10$ p\underline{u}blications ($\fol{U}$). Combination functions are a natural way to compare such {\em relational features} arising from rules. Our network handles this in two steps: first, by ensuring that all instances (papers) of a particular $\fol{Professor-Student}$ pair share the same weights. Second, by combining predictions from each of these instances (papers) using a combination function. We explore the use of {\em Or}, {\em Max} and  {\em Average} combination functions. 
Once the network weights are appropriately constrained by parameter tying and combination functions, they can be learned using standard techniques such as backpropagation. 

We make the following contributions: (1) we learn a NN that can be fully trained from data and with no significant engineering, unlike previous approaches; (2) we combine the successful paradigms of relational random walks and parameter tying from SRL methods; this allows the resulting NN to faithfully model relational data while being fully learnable; (3) we evaluate the proposed approach against recent relational NN approaches and demonstrate its efficacy.

\section{Related Work}

{\bf \bf Lifted Relational Neural Networks.}
Our work is closest to Lifted Relational Neural Networks (LRNN)~\cite{SourekEtAl-16-PredRelNNs} due to \v{S}ourek et al., in terms of the architecture.  LRNN  uses expert hand-crafted relational rules as input, which are then instantiated (based on data) and rolled out as a ground network. 
While at a high-level, our approach appears similar to the LRNN framework, there are significant differences. First, while \v{S}ourek et al., exploit tied parameters across examples within the same rule, there is no parameter tying across multiple instances; our model, however, ensures parameter tying of multiple ground instances of the rule (in our case, a relational random walk). Second, since they adopt a fuzzy notion, their system supports weighted facts (called ground atoms in logic literature). We take a more standard approach and our observations are Boolean. Third, while the previous difference appears to be limiting in our case, note that this leads to a {\em reduction in the number of network weight parameters}.

\^{S}ourek et al., have extended their work to learn network structure using predicate invention~\cite{SourekEtAl-17-StackedRelNNs}; our work learns relational random walks as rules for the network structure. As we show in our experiments, NNs cannot only easily handle such large number of such random walks, but can also use them effectively as a bag of weakly predictive intermediate layers capturing local features. This allows for learning a more robust model than the induced rules, which take a more global view of the domain.
Another recent approach is due to Kazemi and Poole \cite{KazemiPoole18-RelNNs}, who proposed a relational neural network by adding hidden layers to their Relational Logistic Regression \cite{rlr} model. A key limitation of their work is that they are restricted to unary relation predictions, that is,  they can only predict attributes of objects instead of relations between. In contrast, ours is a more general framework in that can be used to predict relations between objects. 

Much of this recent work is closely related to a significant body of research called {\em neural-symbolic integration} \cite{GarcezEtAl02}, which aims to combine (arguably) two of the oldest formalisms in machine learning: symbolic representations with neural learning architectures. Some of the earliest systems such as KBANN \cite{KBANN1990} date back to the early 90s; KBANN also rolls out the network architecture from rules, though it only supports propositional rules. Current work, including ours, instead explores relational rules which serve as templates to roll out more complex architectures. Other recent approaches such as CILP++ \cite{ClipPlusPlus2014} and Deep Relational Machines \cite{DeepRelationalMachines} incorporate relational information as network layers. However, such models  propositionalize relational data into flat-feature vector and hence, cannot be seen as truly relational models. A rather distinctive approach in this vein is due to Hu et al. \cite{TeacherStudentModel2016}, where two independent networks incorporating rules and data are trained together. 
Finally, NNs have also been trained to approximate ILP clause evaluation \cite{DiMaioAndShavlik2004},  
perform SLD-resolution in first-order logic \cite{SLDDeduction2007}, and approximate entailment operators in propositional logic \cite{EntailmentICLR2018}. 


{\bf Relational Random Walks.}
The Path Ranking Algorithm (PRA, \cite{pcrw10}) is a key framework, where a combination of random walks replaces exhaustive search in order to answer queries. Recently, Das et al. \cite{PRARNN2017} considered random walks between query entities to perform composition of embeddings of relations on each walk with recurrent neural networks. 
DeepWalks \cite{Perozzi2014DeepWalkOL} performs random walks on graphs by treating each node as a word, which results in learning embeddings for each node of graph. Kaur et al.\cite{KaurEtAl18-RRBM} consider relational random walks to generate count and existential features to train a relational restricted Boltzmann machine \cite{Larochelle08}. This feature transformation induces propositionalization that could potentially result in loss of information, as we show in our experiments.

{\bf \bf Tensor Based Models.}
Recently, several tensor-based models \cite{RESCAL2011,TRANSE,NTN2013,JointSME2012,TransH2014} have been proposed to learn embeddings of objects and relations. Such models have been very effective for large-scale knowledge-base construction. 
However, they are computationally expensive as they learn parameters for each object and relation in the knowledge base. Furthermore, the embedding into some ambient vector space makes the models more difficult to interpret. Though rule distillation can yield human-readable rules \cite{Yang2015}, it is another computationally intensive post-processing step, which limits the size of the interpreted rules. 

{\bf \bf Other Models.}
Several NNs have been utilized with relational databases schemas \cite{BlockeelRelNN, MINN2000}. These models differ on how they handle 1-to-$N$ joins, cyclicity, and indirect relationships between relations. However, they all learn one network {\em per relation}, which makes them computationally expensive. In the same vein, graph-based models take graph structure into consideration during training. Pham et al. \cite{ColumnNetworks2017} perform collective classification 
via a deep neural network where connections between adjacent layers are established according to given graph structure. Niepert et al. \cite{Niepert2016} proposed an algorithm that prepares the relational data to be directly input to standard convolutional network by assigning an ordering to enable feature convolution. Scarselli et al. \cite{Scarselli2009} proposed Graph Neural Networks in which one neural network is installed at each node of the graph, which is trained by obtaining input from all the incoming edges of graph. One neural network per node makes the model computationally very expensive 
Finally, with the rapid growth of deep learning, relational counterparts of most of existing connectionist models have been also proposed \cite{GraphConvolutionalNetwork2018, RelationalReccurentNetwork2018, RelationalSAE2015, RelConvNN2014}.



%
%

\section{Neural Networks with Relational Parameter Tying}

We first introduce some notation for relational logic, which is used for relational representation, with the domain being represented using constants, variables and predicates. We adopt the following conventions: (1) constants used to represent entities in the domain are written in lower-case (e.g., $\fol{ana}$, $\fol{bob}$); (2) variables and entity types are capitalized (e.g., $\fol{Student}$, $\fol{Professor}$); and (3) relations and predicate symbols between entities and attributes are represented as $\fol{Q(\cdot,  \cdot)}$. A grounding is a predicate applied to a tuple of terms (i.e., either a full or partial instantiation), e.g. $\fol{AdvisedBy(Student, ana)}$, is a partial instantiation. 

Rules are constructed from atoms using logical connectives 
($\wedge$, $\vee$) and quantifiers ($\exists$, $\forall$).  Due to the use of relational random walks, the relational rules that we employ are universally conjunctions of the form $\fol{h \, \Leftarrow \, b_1 \wedge \, \hdots \wedge \, b_\ell}$, where the head $\fol{h}$ is the target of prediction and the body $\fol{b_1 \wedge \, \hdots \wedge \, b_\ell}$ corresponds to conditions that make up the rule (that is, each literal $\fol{b}_i$ in the body is a predicate $\fol{Q(\cdot, \cdot)}$). We do not consider negations in this work.

An example rule could be $\fol{AdvisedBy(S,P)} \, \Leftarrow \, \fol{Professor(P)} \, \wedge \,  \fol{WorksIn(P,T)}  \, \wedge \, \fol{PartOf(T,S)} \, \wedge \, \fol{Student(S)}$. This rules states that if a $\fol{Student}$ is a part of the project that the $\fol{Professor}$ works on, then the $\fol{Student}$ is advised by that $\fol{Professor}$. The body of the rule is learned as a random walk that starts with $\fol{Professor}$ and ends with $\fol{Student}$. Such a random walk represents a chain of relations that could possibly connect a $\fol{Professor}$ to a $\fol{Student}$ and is a {\bf relational feature} that could help in the prediction. The rule head is the target that we are interested in predicting. Since these rules are essentially ``soft" rules, we can also associate clauses with weights, i.e., {\em weighted rules}: $(\fol{R}, w)$.



A relational neural network $\mathcal{N}$ is a set of $M$ weighted rules describing interactions in the domain $\left\{ \fol{R_j}, \, w_j) \right\}_{j = 1}^M$. We are given a set of atomic facts $\F$, known to be true (the evidence) and labeled relational training examples $\{ (\x_i, \, y_i) \}_{i = 1}^\ell$. In general, labels $y_i$ can take multiple values corresponding to a multi-class problem. We seek to learn a relational neural network model $\mathcal{N} \, \equiv \, \left\{ \fol{R_j}, \, w_j) \right\}_{j = 1}^M$ to predict a $\fol{Target}$ relation, given relational examples $\x$, that is: $y \, = \, \fol{Target}(\x)$.\\ 

\noindent \fbox{
\parbox{0.95\textwidth}{
\noindent {\bf Given}: Set of instances $\F$, $\fol{Target}$ relation, relational data set $\left( \x, \, y \right) \in \mathcal{D}$; \\
\noindent {\bf Construct} (structure learning): $\fol{R_j}$, relational random walk rules (relational feature describing the network structure of $\mathcal{N}$); \\
\noindent{\bf Train} (parameter learning): $w_j$, rule weights via gradient descent with rule-based parameter tying to identify a sparse set of network weights of $\mathcal{N}$
}}

$\,$

\begin{example*}
   The movie domain contains the entity types (variables) $\fol{Person}(\fol{P})$, $\fol{Movie}(\fol{M})$ and $\fol{Genre}(\fol{G})$. In addition there are relations (features): $\fol{Directed(P, M)}$, $\fol{ActedIn(P, G)}$ and $\fol{InGenre(M, G)}$.  The domain also has relations for entity resolution: $\fol{SamePerson(P_1,P_2)}$ and $\fol{SameGenre(G_1,G_2)}$. The task is to predict if $\fol{P_1}$  worked under $\fol{P_2}$, with the target predicate (label): $\fol{WorkedUnder(P_1,P_2)}$.
\end{example*}

\subsection{Generating Lifted Random Walks}
The core component of a neural network model is the architecture, which determines how the various neurons are connected to each other, and ultimately how all the input features interact with each other. In a relational neural network, the architecture is determined by the {\em domain structure}, or the set of relational rules that determines how various relations, entities and attributes interact in the domain as shown earlier with the $\fol{AdvisedBy}$ example.  
While previous approaches employed carefully hand-crafted rules, we, instead, use relational random walks to define the network architecture and model the local relational structure of the domain. A similar approach was also used by Kaur et al \cite{KaurEtAl18-RRBM}, though the random walk features were used to instantiate a restricted Boltzmann machine, which has a far more limited architecture and their work is not lifted since it instantiates the entire network before learning. 

Relational data is often represented using a lifted graph, which defines the domain's schema; in such a representation, a relation $\fol{Predicate(Type_1, Type_2)}$ is a predicate edge between two type nodes: $\fol{Type_1 \, \xrightarrow{Predicate} Type_2}$. A relational random walk through a graph is a chain of such edges corresponding to a conjunction of predicates. For a random walk to be semantically sound, we should ensure that the input type (argument domain) of the $(i+1)$-th predicate is the same as the output type (argument range) of the $i$-th predicate. 
\begin{example*}[continued]
The body of the rule
\begin{flalign*}
\fol{ActedIn(P_1, G_1) \wedge SameGenre(G_1, G_2) \wedge ActedIn^{-1}(G_2, P_2) \wedge} & \\ \fol{SamePerson(P_2, P_3) \wedge ActedIn^{-1}(P_3, M) \wedge Directed(M, P_4) }  
& \fol{\Rightarrow WorkedUnder(P_1, P_4)}
\end{flalign*}
can be represented graphically as
\begin{eqnarray*}
\fol{P_1 \xrightarrow{ActedIn} G_1 \xrightarrow{SameGenre} G_2 \xrightarrow{ActedIn^{-1}}} 
\fol{P_2 \xrightarrow{SamePerson} P_3 \xrightarrow{ActedIn^{-1}} M \xrightarrow{Directed} P_4}. 
\end{eqnarray*}
This is a lifted random walk between two entities $\fol{P_1} \rightarrow \fol{P_4}$ in the target predicate, $\fol{WorkedUnder(P_1,P_4)}$. It is semantically sound  as it is possible to chain the second argument of a predicate to the first argument of the succeeding predicate. This walk also contains an {\em inverse predicate} $\fol{ActedIn^{-1}}$, which is distinct from $\fol{ActedIn}$ (since the argument types are reversed).
\end{example*}
We use path-constrained random walks \cite{pcrw10} approach to generate $M$ lifted random walks $\fol{R_j}$, $j = 1,\hdots,M$. These random walks form the backbone of the lifted neural network, as they are templates for various feature combinations in the domain. They can also be interpreted as {\em domain rules} as they impart localized structure to the domain model, that is, they provide a qualitative description of the domain. When these rules, or lifted random walks have weights associated with them, we are then able to endow the rules with a quantitative influence on the target predicate. We now describe a novel approach to network instantiation using these random-walk-based relational features. A key component of the proposed instantiation is rule-based parameter tying, which reduces the number of network parameters to be learned significantly, while still effectively maintaining the quantitative influences as described by the relational random walks.

\subsection{Network Instantiation}
\label{subsec: network instantiation}
\begin{figure*}[t]
\centering
\includegraphics[width=0.8\textwidth]{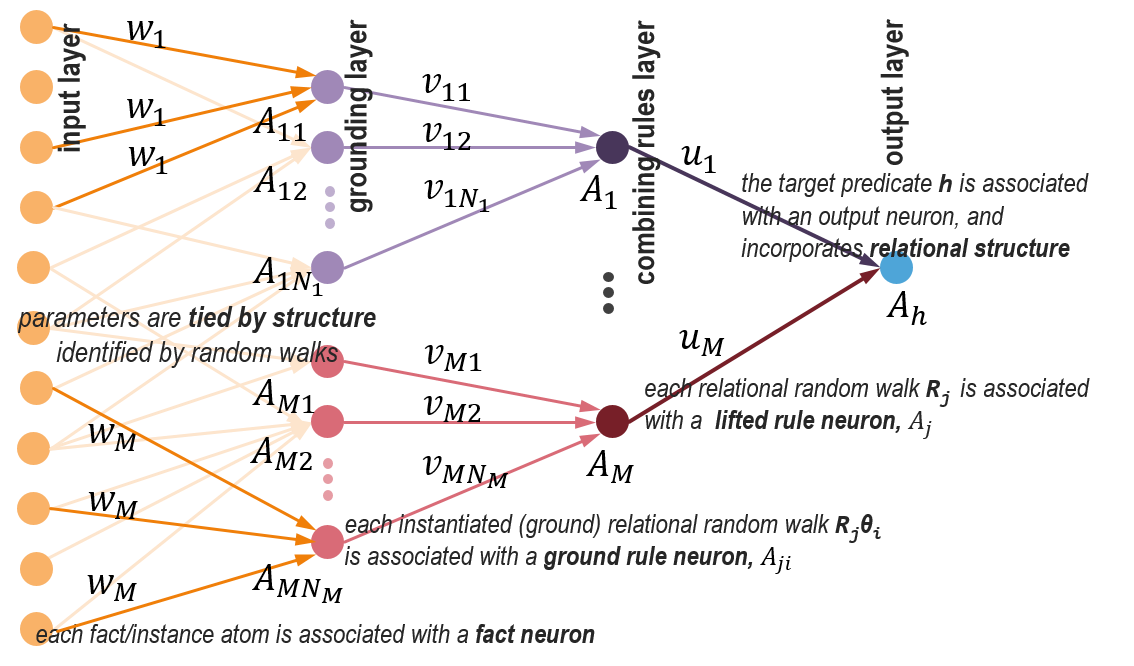}
\vspace{-0.2in}
\caption{
The relational neural network is unrolled in three stages, ensuring that the output is a function of facts through two hidden layers: the combining rules layer (with lifted random walks) and the grounding layer (with instantiated random walks). 
Weights are tied between the input and grounding layers based on which fact/feature ultimately contributes to which rule in the combining rules layer. 
}
\label{fig: unrolling schematic}
\vspace{-0.2in}
\end{figure*}

The relational random walks ($\fol{R_j}$) generated in the previous subsection are the relational features of the lifted relational neural network, $\mathcal{N}$. Our goal is to unroll and ground the network with several intermediate layers that capture the relationships expressed by the random walks. A key difference in network construction between our proposed work and recent approaches such as that of \v{S}ourek et al., \cite{SourekEtAl-15-LRNNs} is that {\em we do not perform an exhaustive grounding} to generate all possible instances before constructing the network. Instead, we only ground as needed leading to a much more compact network. We unroll the network in the following manner (cf. Figure \ref{fig: unrolling schematic}).

{\bf Output Layer}: For the $\fol{Target}$, which is also the head $\fol{h}$ in all the rules $\fol{R_j}$, introduce an output neuron called the {\em target neuron}, $A_\fol{h}$. With one-hot encoding of the target labels, this architecture can handle multi-class problems. The target neuron uses the {\em softmax activation function}. Without loss of generality, we describe the rest of the network unrolling assuming a single output neuron.

{\bf Combining Rules Layer}: The target neuron is connected to $M$ {\em lifted rule  neurons}, each corresponding to one of the lifted relational random walks, $(\fol{R_j}, \, w_j)$. Each rule $\fol{R_j}$ is a conjunction of predicates defined by random walks:
    \begin{equation}
        \fol{Q_1^j(X, \cdot)} \wedge \hdots \wedge \fol{Q_L^j(\cdot, Z)} \, \Rightarrow \, \fol{Target(X, Z)}, \,\, j=1, \hdots, M,
        \label{eq: rule}
    \end{equation}
and corresponds to the lifted rule neuron $A_j$. This layer of neurons is fully connected to the output layer to ensure that all the lifted random walks (that capture the domain structure) influence the output. The extent of their influence is determined by learnable weights, $u_j$ between $A_j$ and the output neuron $A_h$. 
    
In Fig. \ref{fig: unrolling schematic}, we see that the rule neuron $A_j$ is connected to the neurons $A_{ji}$; these neurons correspond to $N_j$ {\em instantiations} of the random-walk $\fol{R_j}$. The lifted rule neuron $A_j$ aims to {\em combine the influence of the groundings/instantiations} of the random-walk feature $\fol{R_j}$ that are true in the evidence. Thus, each lifted rule neuron can also be viewed as a {\em rule combination neuron}. 
The activation function of a rule combination neuron can be any {\em aggregator or combining rule} \cite{NatarajanEtAl08-CRs}. This can include {\em value aggregators} such as {\bf weighted mean, max0}
or {\em distribution aggregators} (if inputs to the this layer are probabilities) such as {\bf Noisy-Or}.
Many such aggregators can be incorporated into the combining rules layer with appropriate weights ($v_{ji}$) and activation functions of the rule neurons.  For instance, 
combining rule instantiations $\textsf{out}(A_{ji})$ with a weighted mean will require learning $v_{ji}$, with the nodes using unit functions for activation. The formulation of this layer is much more general and subsumes the approach of \v{S}ourek et al \cite{SourekEtAl-15-LRNNs}, which uses a max combination layer. 

    
{\bf Grounding Layer}: For each instantiated (ground) random walk $\fol{R_j}\thetab_i, \, i = 1, \hdots, N_j$, we introduce a {\em ground rule neuron}, $A_{ji}$. This ground rule neuron represents the $i$-th instantiation (grounding) of the body of the $j$-th rule, $\fol{R_j \thetab_i}$:  $\fol{Q_1^j \thetab_i \wedge \, \hdots \wedge \, Q_\ell^j \thetab_i}$ (cf. eqn \ref{eq: rule}). The activation function of a ground rule neuron is a {\em logical AND} ($\wedge$); it is only activated when all its constituent inputs are true (that is, only when the entire instantiation is true in the evidence). 

This requires all the constituent facts $\fol{Q_1^j \thetab_i, \, \hdots,  \, Q_\ell^j \thetab_i}$ to be in the evidence. Thus, the $(j, \, i)$-th ground rule neuron is connected to all the {\em fact neurons} that appear in its corresponding instantiated rule body. A key novelty of our approach is regarding relational parameter tying: the weights of connections between the fact and grounding layers are {\em tied} by the rule these facts appear in together. This is described in detail further below.
    
{\bf Input Layer}: Each instantiated (grounded) predicate that appears as a part of an instantiated rule body is a {\em fact}, that is $\fol{Q^j_k \thetab_i} \, \in \F$. For each such instantiated fact, we create a {\em fact neuron} $A_f$, ensuring that each unique fact in evidence has only one single neuron associated with it. Every example is a collection of facts, that is, example $\x_i \equiv \mathcal{F}_i \subset \mathcal{F}$. Thus, an example is input into the system by simply activating its constituent facts in the input layer.

{\bf Relational Parameter Tying}: The most important thing to note about this construction is that we employ {\em rule-based parameter tying} for the weights between the grounding layer and the input/facts layer. Parameter tying ensures that instances corresponding to an example all share the same weight $w_j$ if they occur in the same lifted rule $\fol{R_j}$. The shared weights $w_j$ are propagated through the network in a bottom-up fashion, ensuring that weights in the succeeding hidden layers are influenced by them. 

Our approach to parameter tying is in sharp contrast to that of \v{S}ourek et al., \cite{SourekEtAl-15-LRNNs}, who learn the weights of the network edges between the output layer and the combining rules layer. Furthermore, they also use fuzzy facts (weighted instances), whereas in our case, the facts/instances are Boolean, though their edge weights are tied. Our approach also differs from that of Kaur et al., \cite{KaurEtAl18-RRBM} who also use relational random walks. From a parametric standpoint, Kaur et al., used relational random walks as features for a restricted Boltzmann machine, where the instance neurons and the rule neurons form a bipartite graph. Thus, the relational RBM formulation has significantly more edges, and commensurately many more parameters to optimize during learning.


\begin{figure*}[t]
\includegraphics[width=\textwidth]{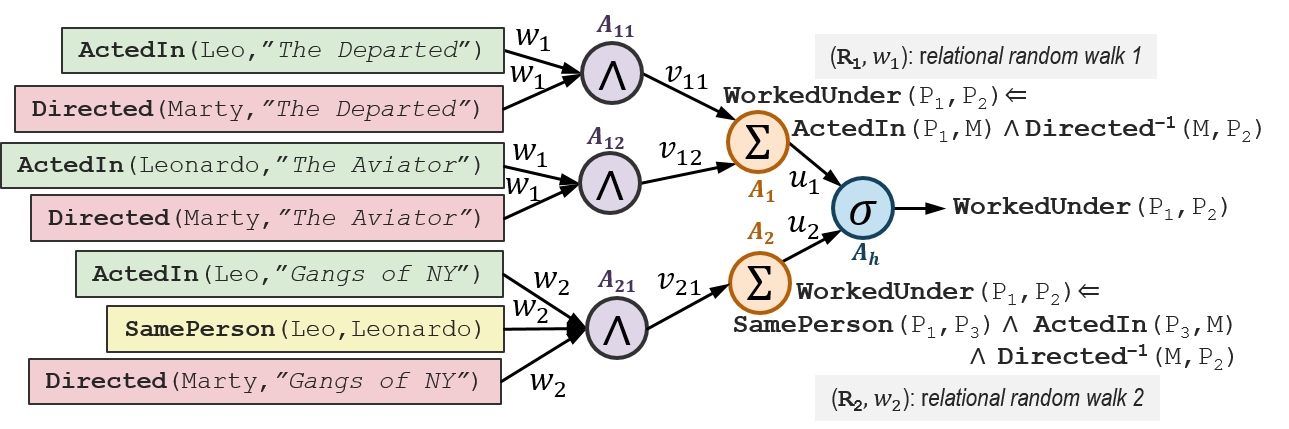}
\vspace{-0.2in}
\caption{Example: unrolling the network with relational parameter tying.}
\label{fig: running example}
\vspace{-0.2in}
\end{figure*}

\begin{example*}[continued, see Fig. \ref{fig: running example}]
Consider two lifted random walks $(\fol{R_1}, w_1)$ and $(\fol{R_2}, w_2)$ for the target predicate $\fol{WorkedUnder(P_1, P_2)}$ 
\begin{flalign*}
\fol{WorkedUnder(P_1, P_2)} \Leftarrow & \fol{ActedIn(P_1, M) \wedge Directed^{-1}(M, P_2)}, \\
\fol{WorkedUnder(P_1, P_2)} \Leftarrow & \fol{SamePerson(P_1, P_3) \wedge ActedIn(P_3, M)   \wedge Directed^{-1}(M, P_2)}.
\end{flalign*}
Note that while the inverse predicate $\fol{Directed^{-1}(M, P)}$ is syntactically different from $\fol{Directed(P, M)}$ (argument order is reversed), they are both semantically same. The {\bf output layer} consists of a single neuron $A_h$ corresponding to the binary target $\fol{WorkedUnder}$. The {\bf lifted rule layer} (also known as combining rules layer) has two lifted rule nodes $A_1$ corresponding to rule $\fol{R_1}$ and $A_2$ corresponding to rule $\fol{R_2}$. These rule nodes combine inputs corresponding to instantiations that are true in the evidence. The network is unrolled based on the specific training example, for instance: $\fol{WorkedUnder(Leo, Marty)}$. For this example, the rule $\fol{R_1}$ has two instantiations that are true in the evidence. Then, we introduce a ground rule node for each such instantiation:
\begin{flalign*}
A_{11}: & \fol{ActedIn(Leo, "The Departed") \wedge Directed^{-1}("The Departed", Marty)}, \\
A_{12}: & \fol{ActedIn(Leo, "The Aviator") \wedge Directed^{-1}("The Aviator", Marty)}.
\end{flalign*}
The rule $\fol{R_2}$ has only one instantiation, and consequently only one node:
\begin{flalign*}
A_{21}: & \fol{SamePerson(Leo, Leonardo) \wedge ActedIn(Leo, "The Departed")} \\ 
 & \wedge \fol{Directed^{-1}("The Departed", Marty)}.
\end{flalign*}
The {\bf grounding layer} consists of ground rule nodes corresponding to instantiations of rules that are true in the evidence. The edges $A_{ji} \rightarrow A_j$ have weights $v_{ji}$ that depend on the combining rule implemented in $A_j$. In this example, the combining rule is {\em average}, so we have $v_{11} = v_{12} = \frac{1}{2}$ and $v_{21} = 1$. The {\bf input layer} consists of atomics fact in evidence: $f \in \mathcal{F}$. The fact nodes $\fol{ActedIn(Leo, "The Aviator")}$ and  $\fol{Directed^{-1}("The Aviator",}$ $\fol{Marty)}$ appear in the grounding $\fol{R_1}\thetab_2$ and are connected to the corresponding ground rule neuron $A_{12}$. Finally, parameters are tied on the edges between the facts layer and the grounding layer. This ensures that all facts that ultimately contribute to a rule are pooled together, which increases the influence of the rule during weight learning. This, in turn, ensures that a rule that holds strongly in the evidence gets a higher weight. 
\end{example*}

Once the network $\mathcal{N}\thetab$ is instantiated, the weights $w_j$ and $u_j$ can be learned using standard techniques such as backpropagation. We denote our approach {\em Neural Networks with Relational Parameter Tying} (\texttt{NNRPT}). The tied parameters incorporate the structure captured by the relational features (lifted random walks), leading to a network with {\em significantly fewer weights}, while also endowing the it with semantic interpretability regarding the discriminative power of the relational features.
We now demonstrate the importance of parameter tying and the use of relational random walks as compared to previous frameworks.

\section{Experiments}
Our empirical evaluation aims to answer the following questions explicitly\footnote{ https://github.com/navdeepkjohal/NNRPT}:
{\bf Q1:}] How does $\mathtt{NNRPT}$ compare to the state-of-the-art SRL models i.e., what the value of learning a neural net over standard models? {\bf Q2:} How does $\mathtt{NNRPT}$ compare to propositionalization models i.e., what is the need for parameterization of standard neural networks? { \bf Q3:} How does $\mathtt{NNRPT}$ compare to other relational neural networks in literature?

\paragraph{Data Sets:}
We use five standard data sets to evaluate our algorithm (see Table \ref{tab: data sets}): 
{\bf \sc Uw-Cse}.
\cite{mln06} is a standard data set that consists of predicates and relations such as $\fol{Professor}$, $\fol{Student}$, $\fol{Publication}$, $\fol{HasPosition}$ and $\fol{TaughtBy}$ etc. The data set contains information from $5$ different areas of computer science about professors, students and courses, and the task is to predict the $\fol{AdvisedBy}$ relationship between a professor and a student. {\bf \sc Imdb} was first created by Mihalkova and Mooney~\cite{bottomupmln07} and contains nine predicates such as $\fol{Gender}$, $\fol{Genre}$, $\fol{Movie}$, 
and $\fol{Director}$. We predict whether an actor has $\fol{WorkedUnder}$ a director. {\bf \sc Cora} is a citation matching data set modified by Poon and Domingos \cite{poon07}. It contains predicates $\fol{Author}$, $\fol{Title}$, $\fol{Venue}$, $\fol{HasWordAuthor}$, $\fol{HasWordTitle}$, $\fol{HasWordVenue}$, $\fol{SameAuthor}$, and $\fol{SameTitle}$. The task is to predict if one venue is $\fol{SameVenue}$ as another.

{\bf \sc Mutagenesis} \cite{lodhi2005} was originally used to predict whether a compound is mutagenetic or not. It consists of properties of compounds, their constituent atoms and the type of bond that exists between atoms. 
We performed relation prediction of whether an atom is a constituent of a given molecule or not ($\fol{MoleAtm(AtomID, MolID)}$). 
{\bf \sc Sports} 
consists of facts from the sports domain crawled by the Never-Ending Language Learner (NELL, \cite{NELL2010}) including details of players, sports, individual plays, league information etc. The goal is to predict which sport a particular team plays.


\begin{table}[!h]
\centering
\setlength{\tabcolsep}{.3\tabcolsep} 
\caption{Data sets used in our experiments to answer {\bf Q1}--{\bf Q3}. The last column shows the number of sampled groundings of random walks per example for $\mathtt{NNRPT}$.}
\begin{tabular}{|c|c|c|c|c|c|c|}
\hline
Domain & Target & \#Facts & \#Pos & \#Neg & \#RW & \#Samp/RW \\
\hline
{\sc Uw-Cse} & $\mathtt{advisedBy}$ & 2817 & 90 & 180 & 2500 & 1000 \\
{\sc Mutagenesis} & $\mathtt{MoleAtm}$ & 29986 & 1000 & 2000 & 100 & 100 \\
{\sc Cora} & $\mathtt{SameVenue}$ & 31086 & 2331 & 4662 & 100 & 100 \\
{\sc Imdb} & $\mathtt{WorkedUnder}$ & 914 & 305 & 710 & 80 & - \\
{\sc Sports} & $\mathtt{TeamPlaysSport}$ & 7824 & 200 & 400 & 200 & 100 \\
\hline
\end{tabular}
\label{tab: data sets}
\vspace{-0.25in}
\end{table}

\paragraph{Baselines and Experimental Details:}
To answer {\bf Q1}, we compare $\mathtt{NNRPT}$ with the more recent and state-of-the-art relational gradient-boosting methods, $\mathtt{RDN}$-$\mathtt{Boost}$\cite{rdnmlj11}, $\mathtt{MLN}$-$\mathtt{Boost}$ \cite{icdm11}, and relational restricted Boltzmann machines $\mathtt{RRBM}$-$\mathtt{E}$, $\mathtt{RRBM}$-$\mathtt{C}$ \cite{KaurEtAl18-RRBM}. As the random walks chain binary predicates in our model, we convert unary and ternary predicates into binary predicates for all data sets. Further, to maintain consistency in experimentation, we use the same resulting predicates across {\em all} our baselines as well. We run $\mathtt{RDN}$-$\mathtt{Boost}$ and $\mathtt{MLN}$-$\mathtt{Boost}$ with their default settings and learn $20$ trees for each model. Also, we train $\mathtt{RRBM}$-$\mathtt{E}$ and $\mathtt{RRBM}$-$\mathtt{C}$ according to the settings recommended in \cite{KaurEtAl18-RRBM}. 

For $\mathtt{NNRPT}$, we generate random walks by considering each predicate and its inverse to be two distinct predicates. Also, we avoid loops in the random walks by enforcing sanity constraints on the random walk generation. We consider $100$ random walks for {\sc Mutagenesis}, {\sc Cora}, $80$ random walks for {\sc Imdb}, $200$ random walks for {\sc Sports} and $2500$ random walks for {\sc Uw-Cse} as suggested by Kaur et al~\cite{KaurEtAl18-RRBM} (see Table \ref{tab: data sets}). Since we use a large number of random walks, {\em exhaustive grounding becomes prohibitively expensive}. To overcome this, we {\em sample groundings} for each random walk for large data sets. Specifically, we sample $100$ groundings per random walk per example for {\sc Cora}, {\sc Sports}, {\sc Mutagenesis}, and $1000$ groundings per random walk per example for {\sc Uw-Cse} (see Table \ref{tab: data sets}). 

For all experiments, we set the positive to negative example ratio to be $1:2$ for training, set combination function to be average and perform $5$-fold cross validation. For $\mathtt{NNRPT}$, we set the learning rate to be $0.05$, batch size to $1$, and number of epochs to $1$. We train our model with $L_1$-regularized AdaGrad \cite{AdaGrad2011}. 
Since these are relational data sets where the data is skewed, AUC-PR and AUC-ROC are better measures than likelihood and accuracy.

To answer {\bf Q2}, we generated flat feature vectors by Bottom Clause Propositionalization (BCP, \cite{ClipPlusPlus2014}), according to which one bottom clause is generated for each example. BCP considers each predicate in the body of the bottom clause as a unique feature when it propositionalizes bottom clauses to flat feature vector. We use Progol \cite{Muggleton1995} to generate these bottom clauses. After propositionalization, we train two connectionist models: a propositionalized restricted Boltzmann machine ($\mathtt{BCP}$-$\mathtt{RBM}$) and a propositionalized neural network ($\mathtt{BCP}$-$\mathtt{NN}$). 
The NN has two hidden layers in our experiments, which makes $\mathtt{BCP}$-$\mathtt{NN}$ model a modified version of CILP++ \cite{ClipPlusPlus2014} that had one hidden layer. The hyper-parameters of both the models were optimized by line search on validation set.

To answer {\bf Q3}, we compare our model with  Lifted Relational Neural Networks ($\mathtt{LRNN}$, \cite{SourekEtAl-15-LRNNs}). To ensure fairness, we perform structure learning by using PROGOL \cite{Muggleton1995} and input the {\em same clauses} to both $\mathtt{LRNN}$ and $\mathtt{NNRPT}$. PROGOL learned $4$ clauses for {\sc Cora}, $8$ clauses for {\sc Imdb}, $3$ clauses for {\sc Sports}, $10$ clauses for {\sc Uw-Cse} and $11$ clauses for {\sc Mutagenesis} in our experiment.

\begin{table*}[t]
 \caption{Comparison of different learning algorithms based on AUC-ROC and AUC-PR. $\mathtt{NNRPT}$ is comparable or better than standard SRL methods across all data sets.}
 \label{tab:tableQ1}
 \centering
\small
\vspace{-0.2in}
\begin{center}
\begin{tabular}{c|p{1.65cm}|p{1.75cm}|p{1.75cm}|p{1.75cm}|p{1.75cm}|p{1.75cm}} 
\Xhline{3\arrayrulewidth}
\Xhline{3\arrayrulewidth}
Data Set & Measure & $\mathtt{RDN}$-$\mathtt{Boost}$ & $\mathtt{MLN}$-$\mathtt{Boost}$ & $\mathtt{RRBM}$-$\mathtt{E}$ & $\mathtt{RRBM}$-$\mathtt{C}$ & $\mathtt{NNRPT}$ \\
\Xhline{3\arrayrulewidth}
\Xhline{3\arrayrulewidth}
\multirow{2}{4em}{\sc Uw-Cse} & AUC-ROC & 
0.973$\pm$0.014&          
0.968$\pm$0.014&          
0.975$\pm$0.013 &         
0.968$\pm$0.011 &         
0.959$\pm$0.024\\[2pt]    
\cline{2-7}
& AUC-PR & 
0.931$\pm$0.036 &         
0.916$\pm$0.035 &         
0.923$\pm$0.056 &         
0.924$\pm$0.040 &         
0.896$\pm$0.063 \\ [2pt]  
\Xhline{3\arrayrulewidth}
\multirow{2}{4em}{\sc Imdb} & AUC-ROC &
0.955$\pm$0.046 &         
0.944$\pm$0.070 &         
1.000$\pm$0.000 &         
0.997$\pm$0.006 &         
0.984$\pm$0.025 \\ [2pt]  
\cline{2-7}
& AUC-PR &
0.863$\pm$0.112&         
0.839$\pm$0.169 &        
1.000$\pm$0.000 &        
0.992$\pm$0.017 &        
0.951$\pm$0.082\\ [2pt]  
\Xhline{3\arrayrulewidth}
\multirow{2}{4em}{\sc Cora} & AUC-ROC &
0.895$\pm$0.183&          
0.835$\pm$0.035 &         
0.984$\pm$0.009 &         
0.867$\pm$0.041 &         
0.952$\pm$0.043  \\ [2pt] 
\cline{2-7}
& AUC-PR &
0.833$\pm$0.259&           
0.799$\pm$0.034 &         
0.948$\pm$0.042 &         
0.825$\pm$0.050 &         
0.899$\pm$0.070 \\ [2pt]  
\Xhline{3\arrayrulewidth}
\multirow{2}{4em}{\sc Mutag.} & AUC-ROC & 
0.999$\pm$0.000&          
0.999$\pm$0.000 &         
0.999$\pm$0.000 &         
0.998$\pm$0.001 &         
0.981$\pm$0.024 \\ [2pt]  
\cline{2-7}
& AUC-PR &
0.999$\pm$0.000 &         
0.999$\pm$0.000 &         
0.999$\pm$0.000 &         
0.997$\pm$0.002 &         
0.970$\pm$0.039 \\ [2pt]  
\Xhline{3\arrayrulewidth}
\multirow{2}{4em}{\sc Sports} & AUC-ROC & 
0.801$\pm$0.026&          
0.806$\pm$0.016 &         
0.760$\pm$0.016 &         
0.656$\pm$0.071 &         
0.780$\pm$0.026  \\ [2pt] 
\cline{2-7}
& AUC-PR &
0.670$\pm$0.028 &        
0.652$\pm$0.032 &        
0.634$\pm$0.020 &        
0.648$\pm$0.085 &        
0.668$\pm$0.070 \\ [2pt] 
\Xhline{3\arrayrulewidth}
\Xhline{3\arrayrulewidth}
\end{tabular}
\end{center}
\vspace{-0.2in}
\end{table*}
 
\begin{table*}[t]
 \caption{Comparison of $\mathtt{NNRPT}$  with propositionalization-based approaches. $\mathtt{NNRPT}$ is significantly better on a majority of data sets.}
 \label{tab:tableQ2}
 \centering
\small
\begin{center}
\begin{tabular}{c|p{1.65cm}|p{1.8cm}|p{1.8cm}|p{1.8cm}} 
\Xhline{3\arrayrulewidth}
\Xhline{3\arrayrulewidth}
Data Set & Measure & $\mathtt{BCP}$-$\mathtt{RBM}$ & $\mathtt{BCP}$-$\mathtt{NN}$ & $\mathtt{NNRPT}$ \\
\Xhline{3\arrayrulewidth}
\Xhline{3\arrayrulewidth}
\multirow{2}{4em}{\sc Uw-Cse} & AUC-ROC & 
0.951$\pm$0.041 &          
0.868$\pm$0.053 &          
0.959$\pm$0.024 \\[2pt]    
\cline{2-5}
& AUC-PR & 
0.860$\pm$0.114 &         
0.869$\pm$0.033 &         
0.896$\pm$0.063 \\ [2pt]  
\Xhline{3\arrayrulewidth}
\multirow{2}{4em}{\sc Imdb} & AUC-ROC & 
0.780$\pm$0.164 &          
0.540$\pm$0.152 &          
0.984$\pm$0.025 \\[2pt]    
\cline{2-5}
& AUC-PR & 
0.367$\pm$0.139 &         
0.536$\pm$0.231 &         
0.951$\pm$0.082 \\ [2pt]  
\Xhline{3\arrayrulewidth}
\multirow{2}{4em}{\sc Cora} & AUC-ROC & 
0.801$\pm$0.017 &          
0.670$\pm$0.064 &          
0.952$\pm$0.043 \\[2pt]    
\cline{2-5}
& AUC-PR & 
0.647$\pm$0.050 &         
0.658$\pm$0.064 &         
0.899$\pm$0.070 \\ [2pt]  
\Xhline{3\arrayrulewidth}
\multirow{2}{4em}{\sc Mutag.} & AUC-ROC & 
0.991$\pm$0.003&           
0.945$\pm$0.019 &          
0.981$\pm$0.024 \\[2pt]    
\cline{2-5}
& AUC-PR & 
0.995$\pm$0.001 &         
0.973$\pm$0.012 &         
0.970$\pm$0.039 \\ [2pt]  

\Xhline{3\arrayrulewidth}
\multirow{2}{4em}{\sc Sports} & AUC-ROC & 
0.664$\pm$0.021 &          
0.543$\pm$0.037 &          
0.780$\pm$0.026 \\[2pt]    
\cline{2-5}
& AUC-PR & 
0.532$\pm$0.041 &         
0.499$\pm$0.065 &         
0.668$\pm$0.070 \\ [2pt]  
\Xhline{3\arrayrulewidth}
\Xhline{3\arrayrulewidth}
\end{tabular}
\end{center}
\vspace{-0.35in}
 \end{table*}

\paragraph{Results:}
Table \ref{tab:tableQ1} compares our $\mathtt{NNRPT}$ to $\mathtt{MLN}$-$\mathtt{Boost}$, $\mathtt{RDN}$-$\mathtt{Boost}$, $\mathtt{RRBM}$-$\mathtt{E}$ and $\mathtt{RRBM}$-$\mathtt{C}$ to answer {\bf Q1}. As we see, $\mathtt{NNRPT}$ is significantly better than $\mathtt{RRBM}$-$\mathtt{C}$ for {\sc Cora} and {\sc Sports} on both AUC-ROC and AUC-PR, and performs comparably to the other data sets. It also performs better than $\mathtt{MLN}$-$\mathtt{Boost}$, $\mathtt{RDN}$-$\mathtt{Boost}$ on {\sc Imdb} and {\sc Cora} data sets, and comparably on other data sets. Similarly, it performs better than $\mathtt{RRBM}$-$\mathtt{E}$ on {\sc Sports}, both on AUC-ROC and AUC-PR and comparably on other data sets. Broadly, {\bf Q1} can be answered affirmatively in that $\mathtt{NNRPT}$ performs comparably to or better than state-of-the-art SRL models.

Table \ref{tab:tableQ2} shows the comparison of $\mathtt{NNRPT}$ with two propositionalization models: $\mathtt{BCP}$-$\mathtt{RBM}$ and $\mathtt{BCP}$-$\mathtt{NN}$ in order to answer {\bf Q2}. $\mathtt{NNRPT}$ performs better than $\mathtt{BCP}$-$\mathtt{RBM}$ on all the data sets except {\sc Mutagenesis}, where the two models have similar performance. $\mathtt{NNRPT}$ also performs better than $\mathtt{BCP}$-$\mathtt{NN}$ on all data sets. It should be noted that BCP feature generation sometimes introduces a large positive-to-negative example skew (for example, in the {\sc Imdb} data set),  which can sometimes gravely affect the performance of the propositional model, as we observe in Table \ref{tab:tableQ2}. 
This emphasizes the need for designing models that can handle relational data directly and without propositionalization; our proposed model as an effort in this direction. {\bf Q2} can now be answered affirmatively: that $\mathtt{NNRPT}$ performs better than propositionalization models.

Table \ref{tab:tableQ3New} compares the performance of  $\mathtt{NNRPT}$ and $\mathtt{LRNN}$ when both use clauses learned by PROGOL \cite{Muggleton1995}. $\mathtt{NNRPT}$ performs better on {\sc Uw-Cse}, {\sc Sports} evaluated using AUC-PR. This result is especially significant because these data sets are considerably skewed. $\mathtt{NNRPT}$ also outperforms $\mathtt{LRNN}$ on {\sc Cora} and {\sc Mutagenesis}. Lastly, $\mathtt{NNRPT}$ has comparable performance on {\sc Imdb} on both AUC-ROC and AUC-PR. The reason for this big performance gap between the two models on {\sc Cora} is likely because $\mathtt{LRNN}$ could not build effective models with the fewer number of clauses (i.e. four) typically learned by PROGOL. In contrast, even with very few clauses, $\mathtt{NNRPT}$ is able to outperform $\mathtt{LRNN}$. This helps us answer {\bf Q3}, affirmatively, that: $\mathtt{NNRPT}$ offers many advantages over state-of-the-art relational neural networks.

In summary, our experiments clearly show the benefits of parameter tying as well as the expressivity of relational random walks in tightly integrating with a neural network model across a wide variety of domains and settings. The key strengths of $\mathtt{NNRPT}$ are that it can (1) efficiently incorporate a large number of relational features, (2) capture local qualitative structure through relational random walk features, (3) tie feature weights (parameter-tying) in a manner that captures the global quantitative influences.

\begin{table*}[t]
 \caption{Comparison of $\mathtt{NNRPT}$ and $\mathtt{LRNN}$ on AUC-ROC and AUC-PR on different data sets. Both the models were provided clauses learnt by PROGOL, \cite{Muggleton1995}. $\mathtt{NNRPT}$ is capable of employing rules to improve performance in some data sets.}
 \label{tab:tableQ3New}
 \centering
\small
\vspace{-0.3in}
\begin{center}
\begin{tabular}{c|p{1.55cm}|p{1.8cm}|p{1.8cm}|p{1.8cm}|p{1.75cm}|p{1.7cm}} 
\Xhline{3\arrayrulewidth}
\Xhline{3\arrayrulewidth}
Model & Measure & {\sc Uw-Cse} & {\sc Imdb} & {\sc Cora} & {\sc Mutagen.} & {\sc Sports} \\
\Xhline{3\arrayrulewidth}
\Xhline{3\arrayrulewidth}

\multirow{2}{4em}{$\mathtt{LRNN}$} & AUC-ROC & 
0.923$\pm$0.027&           %
0.995$\pm$0.004 &
0.503$\pm$0.003 &
0.500$\pm$0.000 &
0.741$\pm$0.016 \\ [2pt]   %
\cline{2-7}
& AUC-PR & 
0.826$\pm$0.056&          %
0.985$\pm$0.013&
0.356$\pm$0.006&
0.335$\pm$0.000&
0.527$\pm$0.036 \\[2pt]   %

\Xhline{3\arrayrulewidth}

\multirow{2}{4em}{$\mathtt{NNRPT}$} & AUC-ROC & 
0.700$\pm$0.186&           %
0.997$\pm$0.007 &
0.968$\pm$0.022 &
0.532$\pm$0.019 &
0.657$\pm$0.014\\ [2pt]   %
\cline{2-7}
& AUC-PR & 
0.910$\pm$0.072&          %
0.992$\pm$0.017 &
0.943$\pm$0.032 &
0.412$\pm$0.032 &
0.658$\pm$0.056 \\[2pt]   %

\Xhline{3\arrayrulewidth}
\Xhline{3\arrayrulewidth}

\end{tabular}
\end{center}
\vspace{-0.3in}
\end{table*}

\paragraph{Discussion:}
A typical convolutional neural network (CNN) is composed of three layers: convolution, max-pooling and (fully-connected) output layers. $\mathtt{NNRPT}$ can be considered a special instance of a convolutional network in relational domains, where the fact-grounding layer edges are the equivalent of convolution, combining rules layer represents pooling, and softmax layer is the fully-connected layer. 
If we perform a full and exhaustive grounding of the neural network in $\mathtt{NNRPT}$, $M$ is the number of lifted random walks (template rules), $N$ is the number of grounded random walks (instances of a template rule) and $|\mathcal{F}|$ is the number of all facts (atomic instances). The data can be represented as a {\em three-dimensional tensor} $B$ of size $M \times N \times |\mathcal{F}|$, whose elements are precisely $B_{ijk} \, = \, \mathbf{Q}_k^j \thetab_i$ (see the discussion of the Input Layer in Section \ref{subsec: network instantiation}).
In addition, if we consider the rule layer as tensor $T$ $=$ $M \times 1 \times |\mathcal{F}|$, where parameters are tied across $|\mathcal{F}|$, then $[w_{m1f}]_{m=1}^{M}$ constitutes the convolving filter that is repeatedly applied to each of $|\mathcal{F}|$ ground instances. The resulting tensor $G = M \times N \times 1$ obtained by composing $G = D \circ T$ representing the output of grounded layer passes through a pooling layer (which is the rule-combination layer, here) to downsample the data produce a new tensor $C = M \times 1 \times 1$. The tensor $C$, when composed with the fully-connected non-linear layer $F = M \times |\mathcal{O}|$ of our model produces tensor of size $1 \times |\mathcal{O}|$ that represents the probability of each class in the output: $\mathcal{O}$.

\section{Conclusion and Future Work}
We considered the problem of learning neural networks from relational data. Our proposed architecture was able to exploit parameter tying i.e., different instances of the same rule shared the same parameters inside the same training example. In addition, we explored the use of relational random walks to create relational features for training these neural nets. 
Further experiments on larger data sets could yield insights into the scalability of this approach. Integration with an approximate-counting method could potentially reduce the training time. Given the relation to CNNs, stacking could allow for our method to be deeper. Finally, understanding the use of such random-walk-based neural network as a function approximator can allow for efficient and interpretable learning in relational domains with minimal feature engineering. \\

\noindent {\bf Acknowledgements}: SN, GK \& NK gratefully acknowledge AFOSR award FA9550-18-1-0462. The authors acknowledge the support of Amazon faculty award. KK acknowledges the support of the  RMU project DeCoDeML. Any opinions, findings, and conclusion or recommendations expressed in this material are those of the authors and do not necessarily reflect the view of the AFOSR, Amazon, DeCoDeML or the US government.

 \bibliographystyle{splncs04}
 \bibliography{RelNNs}
\end{document}